\documentclass{article}

\usepackage{graphicx} 
\usepackage{subfigure} 
\usepackage{amsmath}  
\usepackage{amssymb}  
\usepackage[mathscr]{euscript}
\usepackage{subfigure} 
\usepackage{multirow}

\usepackage{natbib}
\usepackage{wrapfig}

\usepackage{algorithm}
\usepackage{algorithmic}

\newcommand{\mmbf}[1]{\ensuremath{\mathbf{#1}}}
\newcommand{\T}{^\mathsf{T}}

\newcommand{\by}[1]{\ensuremath{#1 \times #1}}

\def\beqa#1\eeqa{\begin{eqnarray}#1\end{eqnarray}}

\def\beqa#1\eeqa{\begin{eqnarray}#1\end{eqnarray}}

\usepackage[accepted]{icml2013}

\icmltitlerunning{Deep Learning using Linear Support Vector Machines}

\begin{document} 

\twocolumn[
\icmltitle{Deep Learning using Linear Support Vector Machines}

\icmlauthor{Yichuan Tang}{tang@cs.toronto.edu}
\icmladdress{Department of Computer Science, University of Toronto. Toronto, Ontario, Canada.}

\icmlkeywords{boring formatting information, machine learning, ICML}

\vskip 0.3in
]

\begin{abstract}
Recently, fully-connected and convolutional neural networks have been trained to achieve state-of-the-art performance on a wide variety of tasks such as speech recognition, image classification, natural language processing, and bioinformatics. For classification tasks, most of these ``deep learning" models employ the softmax activation function for prediction and minimize cross-entropy loss. In this paper, we demonstrate a small but consistent advantage of replacing the softmax layer with a linear support vector machine. Learning minimizes a margin-based loss instead of the cross-entropy loss. While there have been various combinations of neural nets and SVMs in prior art, our results using L2-SVMs show that by simply replacing softmax with linear SVMs gives significant gains on popular deep learning datasets MNIST, CIFAR-10, and the ICML 2013 Representation Learning Workshop's face expression recognition challenge.
\end{abstract}

\section{Introduction}
Deep learning using neural networks have claimed state-of-the-art performances in a wide range of tasks. These include (but not limited to) speech~\cite{MohamedDH09,DahlRMH10} and vision~\cite{JarrettKRL09,CiresanMMGS11,RifaiDVBM11,KrizhevskySH12}. All of the above mentioned papers use the softmax activation function (also known as multinomial logistic regression) for classification.

Support vector machine is an widely used alternative to softmax for classification~\cite{BoserGV92}. Using SVMs (especially linear) in combination with convolutional nets have been proposed in the past as part of a multistage process. In particular, a deep convolutional net is first trained using supervised/unsupervised objectives to learn good invariant hidden latent representations. The corresponding hidden variables of data samples are then treated as input and fed into linear (or kernel) SVMs~\cite{HuangL06, LeeGR09, LeTiled10, CoatesNL11}. This technique usually improves performance but the drawback is that lower level features are not been fine-tuned w.r.t. the SVM's objective.

Other papers have also proposed similar models but with joint training of weights at lower layers using both standard neural nets as well as convolutional neural nets~\cite{ZhongG00, CollobertB04, NagiDGG12}. In other related works,~\citet{WestonR08} proposed a semi-supervised embedding algorithm for deep learning where the hinge loss is combined with the ``contrastive loss" from siamese networks~\cite{Hadsell06}. Lower layer weights are learned using stochastic gradient descent.~\citet{VinyalsNIPS12} learns a recursive representation using linear SVMs at every layer, but without joint fine-tuning of the hidden representation.

In this paper, we show that for some deep architectures, a linear SVM top layer instead of a softmax is beneficial. We optimize the primal problem of the SVM and the gradients can be backpropagated to learn lower level features. Our models are essentially same as the ones proposed in~\cite{ZhongG00, NagiDGG12}, with the minor novelty of using the loss from the L2-SVM instead of the standard hinge loss. Unlike the hinge loss of a standard SVM, the loss for the L2-SVM is differentiable and penalizes errors much heavily. The primal L2-SVM objective was proposed 3 years before the invention of SVMs~\cite{Hinton89}! A similar objective and its optimization are also discussed by~\cite{LeeM01}.

Compared to nets using a top layer softmax, we demonstrate superior performance on MNIST, CIFAR-10, and on a recent Kaggle competition on recognizing face expressions. Optimization is done using stochastic gradient descent on small minibatches. Comparing the two models in Sec.~\ref{sec:regu}, we believe the performance gain is largely due to the superior regularization effects of the SVM loss function, rather than an advantage from better parameter optimization.

\section{The model}
\subsection{Softmax}
For classification problems using deep learning techniques, it is standard to use the softmax or 1-of-K encoding at the top. For example, given 10 possible classes, the softmax layer has 10 nodes denoted by $ p_i $, where $ i=1, \dots, 10 $. $ p_i $ specifies a discrete probability distribution, therefore, $ \sum_i^{10} p_i = 1 $. 

Let $ \mmbf{h} $ be the activation of the penultimate layer nodes, $ \mmbf{W} $ 
is the weight connecting the penultimate layer to the softmax layer, the total input into a softmax layer, given by $ \mmbf{a} $, is
\begin{align}
a_i &= \sum_k h_k W_{ki},
\end{align}
then we have
\begin{align}
p_i &= \frac{\exp(a_i)}{\sum_j^{10} \exp(a_j) }
\end{align}
The predicted class $ \hat{i} $ would be
\begin{align} \label{eq:softmaxpred}
\hat{i} &=  \operatorname*{arg\,max}_{i} p_i  \nonumber \\
 &=  \operatorname*{arg\,max}_{i} a_i
\end{align}

\subsection{Support Vector Machines}
Linear support vector machines (SVM) is originally formulated for binary classification. Given training data and its corresponding labels $ (\mathbf{x}_n, y_n) $, $ n=1, \dots, N $, $ \mathbf{x}_n \in \mathbb{R}^D $, $ t_n \in \{ -1, +1 \} $, SVMs learning consists of the following constrained optimization:
\begin{align}
\min_{\mathbf{w}, \xi_n} & \ \ \frac{1}{2} \mathbf{w} \T \mathbf{w} + C \sum_{n=1}^N \xi_n \\
s.t. & \ \  \mathbf{w} \T \mathbf{x}_n t_n \geq 1 - \xi_n \ \ \ \forall n  \nonumber \\
& \ \ \xi_n \geq 0 \ \ \ \forall n \nonumber
\end{align}
$ \xi_n $ are slack variables which penalizes data points which violate the margin requirements. Note that we can include the bias by augment all data vectors $ \mathbf{x}_n $ with a scalar value of $ 1 $. The corresponding unconstrained optimization problem is the following:
\begin{equation}\label{eq:l1svm}
\min_{\mathbf{w}} \ \ \frac{1}{2} \mathbf{w} \T \mathbf{w} + C \sum_{n=1}^N \max(1-\mathbf{w} \T \mathbf{x}_n t_n, 0)
\end{equation}
The objective of Eq.~\ref{eq:l1svm} is known as the primal form problem of L1-SVM, with the standard hinge loss. Since
L1-SVM is not differentiable, a popular variation is known as the L2-SVM which minimizes the squared hinge loss:
\begin{equation}\label{eq:l2svm}
\min_{\mathbf{w}} \ \ \frac{1}{2} \mathbf{w} \T \mathbf{w} + C \sum_{n=1}^N \max(1-\mathbf{w} \T \mathbf{x}_n t_n, 0)^2
\end{equation}
L2-SVM is differentiable and imposes a bigger (quadratic vs. linear) loss for points which violate the margin.
To predict the class label of a test data $ \mathbf{x} $:
\begin{align}
\operatorname*{arg\,max}_{t} (\mathbf{w} \T \mathbf{x}) t
\end{align}

For Kernel SVMs, optimization must be performed in the dual. However, scalability is a problem with Kernel SVMs, and in this paper we will be only using linear SVMs with standard deep learning models.

\subsection{Multiclass SVMs}
The simplest way to extend SVMs for multiclass problems is using the so-called \emph{one-vs-rest} approach~\cite{Vapnik95}. For $ K $ class problems, $ K $ linear SVMs will be trained independently, where the data from the other classes form the negative cases. ~\citet{HsuL02} discusses other alternative multiclass SVM approaches, but we leave those to future work.

Denoting the output of the $ k $-th SVM as 
\begin{equation}
a_k(\mathbf{x}) = \mathbf{w} \T \mathbf{x}
\end{equation}
The predicted class is
\begin{equation} \label{svmpredict}
\operatorname*{arg\,max}_{k} a_k(\mathbf{x})
\end{equation}
Note that prediction using SVMs is exactly the same as using a softmax Eq.~\ref{eq:softmaxpred}. The only difference between softmax and multiclass SVMs is in their objectives parametrized by all of the weight matrices $ \mathbf{W} $. Softmax layer minimizes cross-entropy or maximizes the log-likelihood, while SVMs simply try to find the maximum margin between data points of different classes.

\subsection{Deep Learning with Support Vector Machines}
Most deep learning methods for classification using fully connected layers and convolutional layers have used softmax layer objective to learn the lower level parameters. There are exceptions, notably in papers by~\cite{ZhongG00, CollobertB04, NagiDGG12}, supervised embedding with nonlinear NCA~\cite{SalakH07}, and semi-supervised deep embedding~\cite{WestonR08}. In this paper, we use
L2-SVM's objective to train deep neural nets for classification. Lower layer weights are learned by backpropagating the gradients from the top layer linear SVM. To do this, we need to differentiate the SVM objective with respect to the activation of the penultimate layer. Let the objective in Eq.~\ref{eq:l1svm} be $ l(\mathbf{w}) $, and the input $ \mathbf{x} $ is replaced with the penultimate activation $ \mathbf{h} $,
\begin{equation}
\frac{\partial l(\mathbf{w}) }{\partial \mathbf{h}_n } =  - C t_n \mathbf{w}(\mathbb{I} \{ 1 > \mathbf{w} \T \mathbf{h}_n t_n\})
\end{equation}
Where $ \mathbb{I}\{\cdot\} $ is the indicator function. Likewise, for the L2-SVM, we have
\begin{equation}
\frac{\partial l(\mathbf{w}) }{\partial \mathbf{h}_n } =  - 2C t_n \mathbf{w} \big( \max (1-\mathbf{w} \T \mathbf{h}_n t_n, 0) \big )
\end{equation}
From this point on, backpropagation algorithm is exactly the same as the standard softmax-based deep learning networks. We found L2-SVM to be slightly better than L1-SVM most of the time and will use the L2-SVM in the experiments section.

\section{Experiments}
\subsection{Facial Expression Recognition}
This competition/challenge was hosted by the ICML 2013 workshop on representation learning, organized by 
the LISA at University of Montreal. The contest itself was hosted on Kaggle with over 120 competing teams
during the initial developmental period.

\begin{figure}[t]
  \begin{center}
    \includegraphics[width=0.45\textwidth]{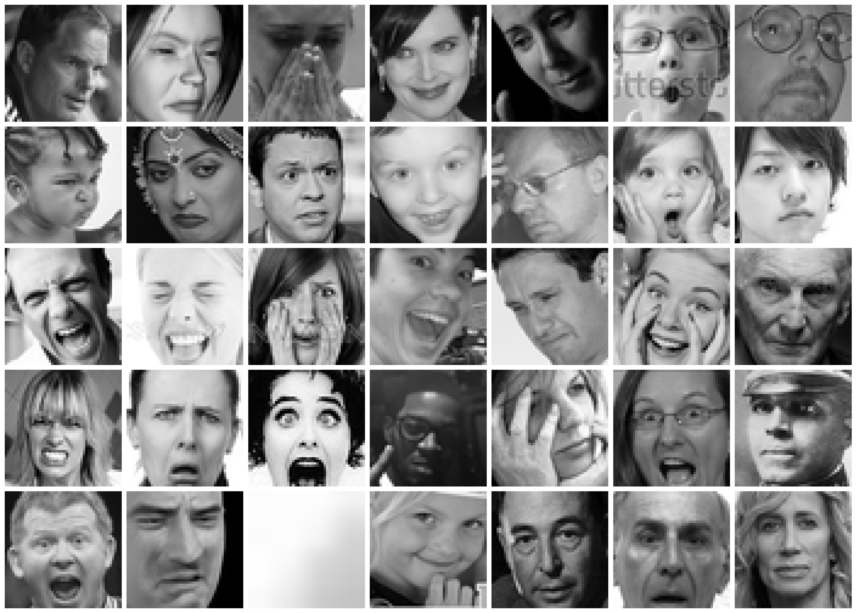}
  \end{center}
  \caption{\textit{Training data. Each column consists of faces of the same expression: starting from the 
  leftmost column: Angry, Disgust, Fear, Happy, Sad, Surprise, Neutral.}}
  \label{fig:faceexp}  
\end{figure}

The data consist of 28,709 48x48 images of faces under 7 different types of expression. See Fig~\ref{fig:faceexp}
for examples and their corresponding expression category. The validation and test sets consist of 3,589 
images and this is a classification task.

\subsubsection*{Winning Solution}
\label{sec:kaggle_winning}
We submitted the winning solution with a public validation score of 69.4\% and corresponding private test
score of 71.2\%. Our private test score is almost 2\% higher than the 2nd place team. Due to label noise and
other factors such as corrupted data, human performance is roughly estimated to be between 65\% and 68\%\footnote{Personal communication from the competition organizers: http://bit.ly/13Zr6Gs}.

Our submission consists of using a simple Convolutional Neural Network with linear one-vs-all SVM at the top. Stochastic gradient descent with momentum is used for training and several models are averaged to slightly improve the generalization capabilities. Data preprocessing consisted of first subtracting the mean value of each image and then setting the image norm to be 100. Each pixels is then standardized by removing its mean and dividing its value by the standard deviation of that pixel, across all training images.

Our implementation is in C++ and CUDA, with ports to Matlab using MEX files. Our convolution routines used fast CUDA kernels written by Alex Krizhevsky\footnote{http://code.google.com/p/cuda-convnet}.
The exact model parameters and code is provided on by the author at https://code.google.com/p/deep-learning-faces.

\subsubsection{Softmax vs. DLSVM}
We compared performances of softmax with the deep learning using L2-SVMs (DLSVM). Both models are tested using an 8 split/fold cross validation, with a image mirroring layer, similarity transformation layer, two convolutional filtering+pooling stages, followed by a fully connected layer with 3072 hidden penultimate hidden units. The hidden layers are all of the rectified linear type. other hyperparameters such as weight decay are selected using cross validation.

\begin{table}[h]
\begin{tabular}{c|c|c}
  \hline
  		 & Softmax & DLSVM L2  \\ 
  	\hline   
    Training cross validation & 67.6\%  & 68.9\% \\
  	\hline   
    Public leaderboard &  69.3\% &  69.4\% \\
	\hline
	Private leaderboard & 70.1\% & 71.2\% \\
	\hline
\end{tabular}
\caption{Comparisons of the models in terms of \% accuracy. Training c.v. is the average cross validation accuracy over 8 splits. Public leaderboard is the held-out validation set scored via Kaggle's public leaderboard. Private leaderboard is the final private leaderboard score used to determine the competition's winners.}
\end{table}

\begin{figure}[h]
  \begin{center}
    \includegraphics[width=0.45\textwidth]{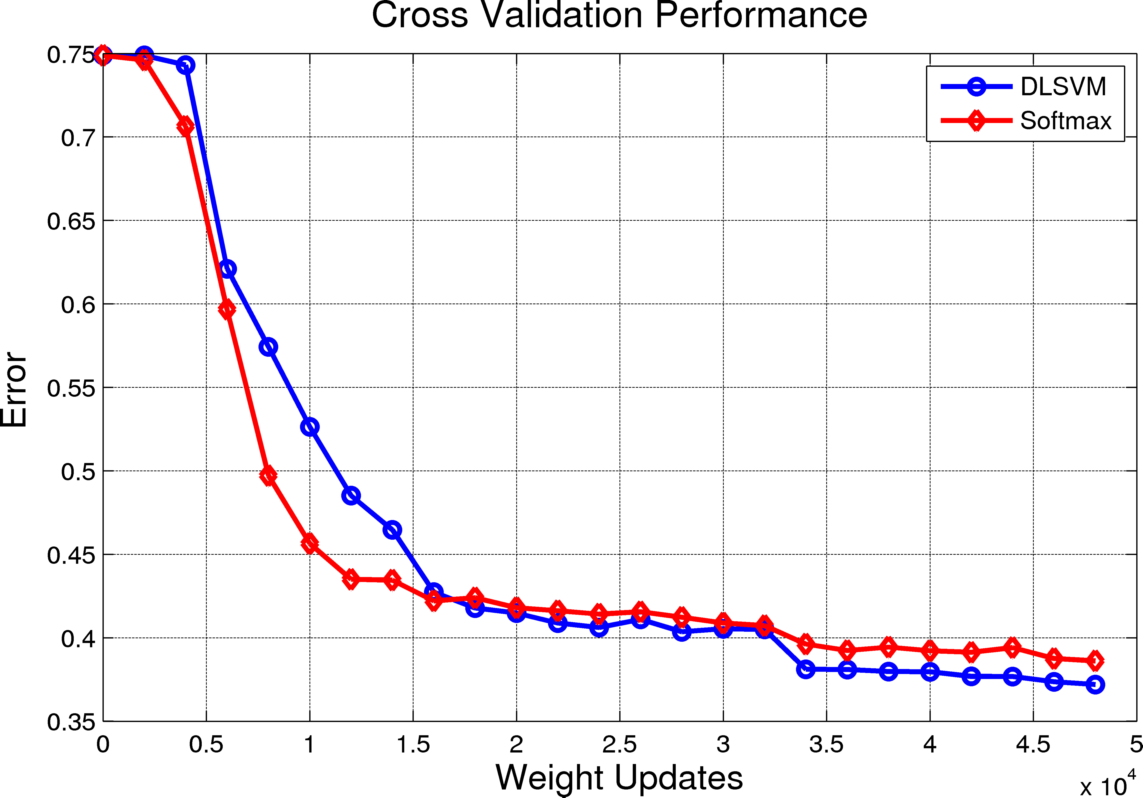}
  \end{center}
  \caption{\textit{Cross validation performance of the two models. Result is averaged over 8 folds.}}
  \label{fig:facecurve}  
\end{figure}

We can also look at the validation curve of the Softmax vs L2-SVMs as a function of weight updates in Fig.~\ref{fig:facecurve}.
As learning rate is lowered during the latter half of training, DLSVM maintains a small yet clear performance gain. 

We also plotted the 1st layer convolutional filters of the two models:
\begin{figure}[h]
  \begin{center}
    \includegraphics[width=0.4\textwidth]{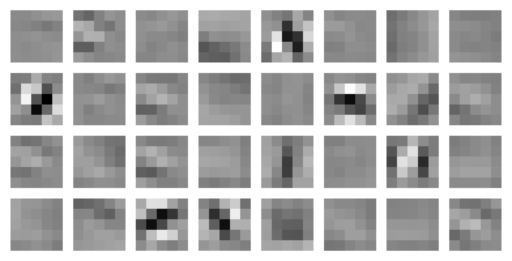}
  \end{center}
  \caption{\textit{Filters from convolutional net with softmax.}}
  \label{fig:softmaxfilters}  
\end{figure}

\begin{figure}[h]
  \begin{center}
    \includegraphics[width=0.4\textwidth]{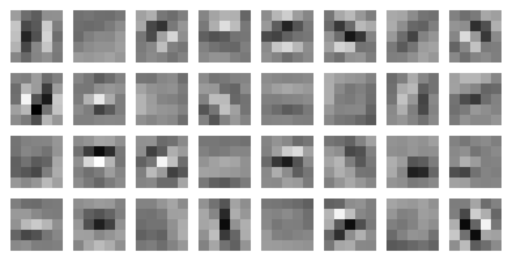}
  \end{center}
  \caption{\textit{Filters from convolutional net with L2-SVM.}}
  \label{fig:svm_filters}  
\end{figure}
While not much can be gain from looking at these filters, SVM trained conv net appears to have more textured filters.

\subsection{MNIST}
MNIST is a standard handwritten digit classification dataset and has been widely used as a benchmark dataset in deep learning. It is a 10 class classification problem with 60,000 training examples and 10,000 test cases.

We used a simple fully connected model by first performing PCA from 784 dimensions down to 70 dimensions. Two hidden layers of 512 units each is followed by a softmax or a L2-SVM. The data is then divided up into 300 minibatches of 200 samples each. We trained using stochastic gradient descent with momentum on these 300 minibatches for over 400 epochs, totaling 120K weight updates. Learning rate is 
linearly decayed from 0.1 to 0.0. The L2 weight cost on the softmax layer is set to 0.001. To prevent overfitting and critical to achieving good results, a lot of Gaussian noise is added to the input. Noise of standard deviation of 1.0 (linearly decayed to 0) is added. The idea of adding Gaussian noise is taken from these papers~\cite{RaikoVL12,RifaiGBV11}.

Our learning algorithm is permutation invariant without any unsupervised pretraining and obtains these results:
\textbf{\textit{
Softmax: 0.99\% \ \ \ \
DLSVM: 0.87\%
}}

An error of 0.87\% on MNIST is probably (at this time) state-of-the-art for the above learning setting.
The only difference between softmax and DLSVM is the last layer. This experiment is mainly to demonstrate the effectiveness of the last linear SVM layer vs. the softmax, we have not exhaustively explored other commonly used tricks such as Dropout, weight constraints, hidden unit sparsity, adding more hidden layers and increasing the layer size.

\subsection{CIFAR-10}
Canadian Institute For Advanced Research 10 dataset is a 10 class object dataset with 50,000 images for training and 10,000 for testing. The colored images are \by{32} in resolution.
We trained a Convolutional Neural Net with two alternating pooling and filtering layers. Horizontal reflection and jitter is applied to the data randomly before the weight is updated using a minibatch of 128 data cases. 

The Convolutional Net part of both the model is fairly standard, the first C layer had 32 \by{5} filters with Relu hidden units, the second C layer has 64 \by{5} filters. Both pooling layers used max pooling and downsampled by a factor of 2.

The penultimate layer has 3072 hidden nodes and uses Relu activation with a dropout rate of 0.2. The difference between the Convnet+Softmax and ConvNet with L2-SVM is the mainly in the SVM's C constant,
the Softmax's weight decay constant, and the learning rate. We selected the values of these hyperparameters for each model separately using validation.

\begin{table}[h]
\centering
\begin{tabular}{c|c|c}
  \hline
  		 		& ConvNet+Softmax & ConvNet+SVM  \\ 
  	\hline   
    Test error  & 14.0\%  & 11.9\% \\
  	\hline  
\end{tabular}
\caption{Comparisons of the models in terms of \% error on the test set.}
\end{table}
In literature, the state-of-the-art (at the time of writing) result is around 9.5\% by (Snoeck et al. 2012). However, that model is different as it includes contrast normalization layers as well as used Bayesian optimization to tune its hyperparameters.

\subsection{Regularization or Optimization}\label{sec:regu}
To see whether the gain in DLSVM is due to the superiority of the objective function or
to the ability to better optimize, We looked at the two final models' loss under its own objective functions as well as the other objective. The results are in Table~\ref{tab:ro}.

\begin{table}[h]
\centering
\begin{tabular}{c|c|c}
  \hline
  		 		& ConvNet & ConvNet  \\ 
  		 		& +Softmax & +SVM  \\ 
  	\hline   
    Test error  	& 14.0\%  & 11.9\% \\
  	\hline
  	Avg. cross entropy   &  0.072 &  0.353 \\
  	\hline
  	Hinge loss squared  & 213.2 & 0.313 \\
  	\hline  
\end{tabular}
\caption{Training objective including the weight costs.}
\label{tab:ro}
\end{table}

It is interesting to note here that lower cross entropy actually led a higher error in the middle row. In addition, we also initialized a ConvNet+Softmax model with the weights of the DLSVM that had 11.9\% error.
As further training is performed, the network's error rate gradually increased towards 14\%.

This gives limited evidence that the gain of DLSVM is largely due to a better objective function.

\section{Conclusions}\vspace*{-0.05in}
In conclusion, we have shown that DLSVM works better than softmax on 2 standard datasets and a recent dataset. Switching from softmax to SVMs is incredibly simple and appears to be useful for classification tasks. Further research is needed to explore other multiclass SVM formulations and better understand where and how much the gain is obtained.

\subsection*{Acknowledgment}
Thanks to Alex Krizhevsky for making his very fast CUDA Conv kernels available!
Many thanks to Relu Patrascu for making running experiments possible!
Thanks to Ian Goodfellow, Dumitru Erhan, and Yoshua Bengio for organizing the contests.


\begin{small}
\bibliography{DeepBelief,vision}
\bibliographystyle{icml2013}
\end{small}

\end{document}